\documentclass[10pt, a4paper]{article}

\usepackage{booktabs} 
\usepackage[compact]{titlesec} 
\titlespacing{\section}{0pt}{0pt}{0pt} 

\AtBeginDocument{%
  \setlength\abovedisplayskip{0pt}
  \setlength\belowdisplayskip{0pt}}

\usepackage{lrec-coling2024} 

\usepackage{natbib} 
\usepackage{graphicx} 
\usepackage{tabularx} 
\usepackage{soul} 
\usepackage{xcolor} 
\usepackage{hyperref} 
\usepackage{ifthen} 
\usepackage{fontenc} 
\usepackage{microtype} 
\usepackage{amsmath} 
\usepackage{amsfonts} 
\usepackage{amssymb} 

\NewDocumentCommand{\note}{ mO{} }{\textcolor{red}{\textsuperscript{\textit{Note:}}\textsf{\textbf{\small#1}}}}

\title{SCI 3.0: A Web-based Schema Curation Interface for Graphical Event Representations}

\name{Reece Suchocki, Mary Martin, Martha Palmer, Susan Brown} 

\address{The University of Colorado, Boulder \\
         Reece.Suchocki@colorado.edu, Mary.Martin@colorado.edu, \\Martha.Palmer@colorado.edu, Susan.Brown@colorado.edu\\}

\abstract{
To understand the complexity of global events, one must navigate a web of interwoven sub-events, identifying those most impactful elements within the larger, abstract macro-event framework at play.
This concept can be extended to the field of natural language processing (NLP) 
through the creation of structured event schemas which can serve as representations of these abstract events. Central to our approach is the Schema Curation Interface 3.0 (SCI 3.0), a web application that facilitates real-time editing of event schema properties within a generated graph e.g., adding, removing, or editing sub-events, entities, and relations directly through an interface. 
\ \newline  }

\begin{document}

\maketitleabstract

\section{Introduction}

Events, whether reported in the media or through natural discourse, primarily embody properties of hierarchy, sequence, and relation. These properties carry key information as to the scope and significance of an event occurrence. 

While the bulk of this information might be available in a given news article, each real event contributes to a conceptual understanding of an event prototype, which forms the basis for the perception of what constitutes normalcy.
An empirical schematic event representation aids event analysts to perceive prototypical event types free of their already perceived constructs. 

By their nature, events are characterized by distinct start and end points, never occurring in absolute isolation. Instead, they manifest and recur across different contexts, featuring varying participants and consequences. This recurrence allows for the abstraction and quantification of patterns across entire event domains. In the context of this research, event schemas consist of three major components: events, entities, and relations. Events operate within a hierarchical framework, with key actors, embodied as entities, making appearances in multiple events when necessary as participants. The connections between entities, denoted as entity-entity relations, bind these actors together as they engage in the temporal sequence of an event.
Event schemas serve the purpose of defining a typical progression of sub-events, supporting tasks such as information extraction, knowledge base construction, and event prediction. Our approach to modeling a comprehensive schema library involves several steps. First, we establish a format for representing schematic events, which serves as the foundation for our work. Next, we use an automatic induction process to generate an initial library of event schemas. This induction process is based on analyzing a collection of news articles, allowing us to capture a wide range of real-world events. However, it is important to note that this initial library almost always requires further refinement. We employ manual curation to improve the schema library's quality and coverage. This is where the Schema Curation Interface (SCI) 3.0 comes into play. The SCI serves as a powerful tool that facilitates the manual curation of an event schema library. Analysts can utilize the SCI to incorporate human-suggested feedback and make necessary adjustments to the schema library.



\section{Existing Research}

In the field of natural language processing (NLP), much research has been done in the area of extending simple events into more complex event-entity relations. This has resulted in work such as ISO-TimeML \citep{pustejovsky2010iso} and Richer Event Descriptions \citep{o2021richer}. Both papers investigate how annotation can elucidate complex relations, and how they contribute to a schema data format able to be read by a computer.

Graph visualization tools for treebanks, PropBanks \citep{akbik2017projector}, and ontologies \citep{ghorbel2016memo} are already found in practice. To our knowledge, the SCI 3.0 is among the first open-source interfaces for complex event schema visualization and editing.

With the advent of large language modeling \citep{li2020connecting} and graph modeling \citep{li2021future}, automatically induced event schemas have emerged as a promising field of research. However, automatically induced schemas are often noisy, have limited coverage, and are unsuitable for downstream tasks \citep{du2022resin}. This innovation has created a need for human schema revision and curation tools. 

\section{Automatic Schema Induction}\label{sec:auto_induction}
The starting point for schema curation is an automatically generated schema graph. We use the system described in \cite{hierarchicalschema2023}, which can produce hierarchical schemas with the power of OpenAI's GPT-3 model text-davinci-003. The input to the schema induction algorithm is the scenario name, a collection of relevant news articles (from Wikipedia), and a chapter structure. This chapter structure serves as the top level of the hierarchy.
The algorithm consists of three rounds: (1) event skeleton construction; (2) event expansion and (3) event-event relation verification. The first round will produce a linear chain of events that belong to the chapter. Then, for each of those events, we run the event expansion step to obtain its neighbor events: events that are connected through temporal or hierarchical edges. Finally, for each pair of events in the same chapter, we run the event-event relation verification step to compute the confidence of the edge between the pair and check for global consistency. 

One caveat of this algorithm is that the events are represented in the form of short sentences. To ground these events within the ontology, allowing us to extract event arguments, we employ a natural language inference (NLI) model to compare the generated sentences with the definitions of the QNodes. However, even after this grounding procedure, we still encounter challenges related to entity coreference and entity-entity relations, which require manual intervention by curators.

\section{Schema Curation}

SCI 1.0 was developed to generate a visual representation of complex events through a standardized data input, predating our described Schema Data Format (SDF) 
\citep{mishra2021graphical}. The interface supported editing, but was restricted to representing a sequence of primitive events within a one-dimensional timeline. Functionally, it was limited to a generalization without hierarchical event structures, and visually, the graph could quickly become crowded. However, these constraints have been addressed in the latest iteration of our interface. 

SCI 3.0 is a web application designed to support the curation of large event libraries. Its primary objective is to serve as a tool for analysts interested in modeling complex events, entities, and their relations. The current version of the tool now accommodates hierarchical event structures, logic gates, temporal sequences, and a thorough user interface overhaul. This includes the option of direct schema editing via the graph, circumventing the need for any editing directly within the JSON-based SDF, as was necessary in SCI 1.0.

\subsection{Overview of SCI 3.0}

\begin{figure}
    \centering
    \includegraphics[width=\linewidth]{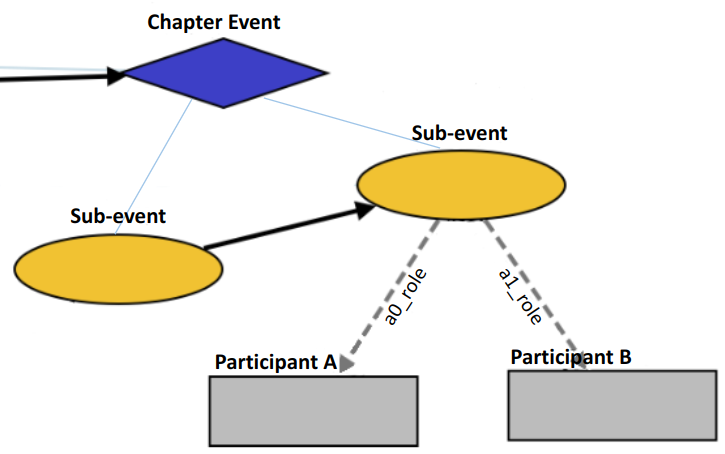}
    \caption{Snapshot of the event schema graph in the SCI Viewer. Sub-events are linked as children of the chapter events and feature participants, with their roles described by the argument labels along the dashed directed edges.}
    \label{fig:graph-example}
\end{figure}

SCI 3.0 is a full-stack web interface that enables users to intuitively upload complex event schemas and visualize the hierarchical structure of event-entity relations. React.js, Flask, and Cytoscape.js were used to design the web application. React is a JavaScript library for fast UI feature integration, Flask is a Python web framework for server communication, and Cytoscape is a JavaScript graph library for interactive graph display and editing \citep{franz2016cytoscape}. Our pipeline works with a React front-end interface which passes the uploaded schema to the Flask app in the back-end to make changes in the JSON schema file. The graph is then constructed with nodes and edges in Cytoscape for rendering on the browser. Major components of the tool now include a retractable JSON editor, an extensive right-click context menu on the graph, and a much larger workspace for human curation. Figure \ref{fig:graph-example} displays a section of the rendered event schema graph, along with a sample layout illustrating chapter and sub-event nodes with participants.


\begin{figure}[h]
    \includegraphics[width=0.3\textwidth]{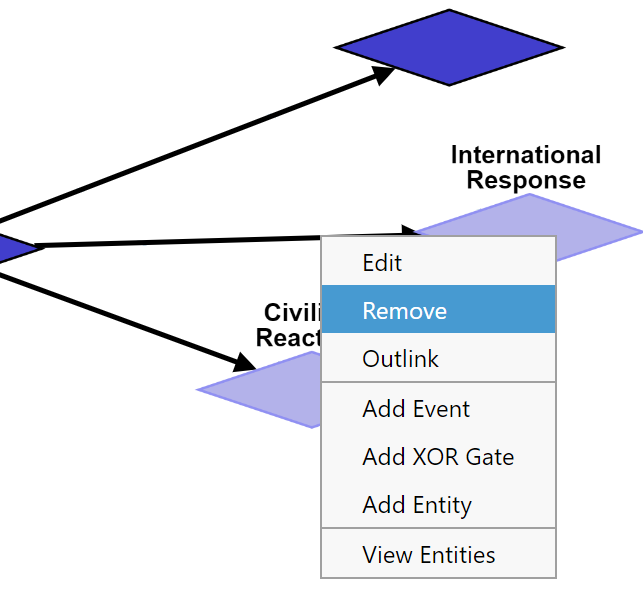}
    \caption{Options displayed when any chapter or primitive event node is right-clicked.}
    \label{context_menu}
\end{figure}

SCI 3.0 allows users to add and edit elements in a schema graph easily. To add events, users can right-click on any node in the graph and select the \textit{Add event} option [\ref{context_menu}]. A window will appear, providing fields for the event's label, associated QNode, description, and a switch to choose between a chapter event or a primitive event.

\begin{figure}[h]
    \centering
    \includegraphics[width=\linewidth]{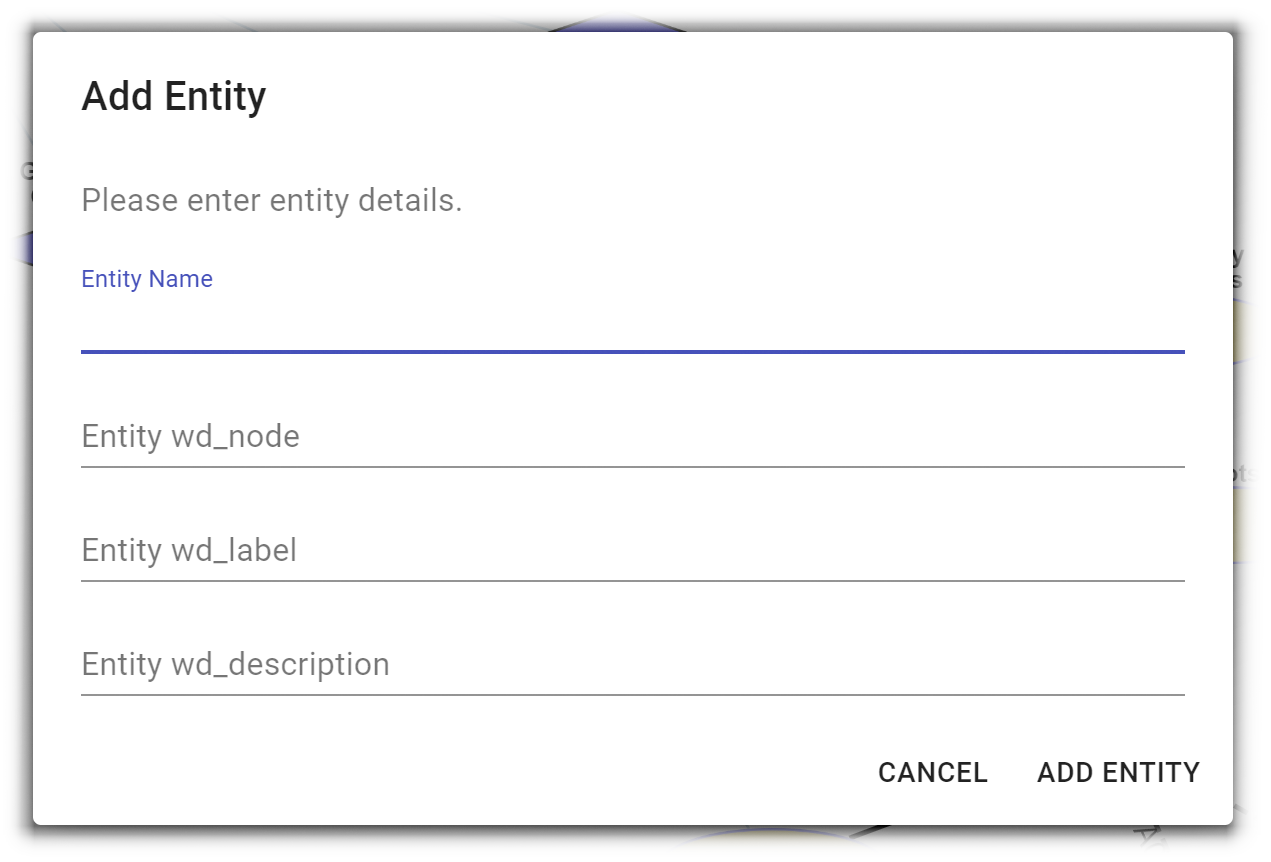}
    \caption{Prompt for creating an entity, along with its required fields.}
    \label{add_entity}
\end{figure}

\begin{itemize}
    \item \textbf{Participants} can be added to a primitive event by right-clicking the event node and selecting \textit{add participant}. Users will then be prompted to provide the semantic role of the participant, the participant name, and the existing entity that is being assigned to the role. A drop-down menu within the entity field provides a list of preexisting entities from which an entity can be selected.
    \item To add a \textbf{relation}, users can right-click on a participant and choose the \textit{add relation} option. This action prompts the user to input the associated name, QNode, WikiData label, description, and the target entity of the relation. The resulting relation is rendered as a directed edge between the selected entities. This edge is labeled with the corresponding relation name.
    \item \textbf{Outlinks} indicate the temporal relationships among events within a schema. To add an outlink, users can right-click on any node in the graph and select the \textit{outlink} option. By then selecting another node of the same event type, an outlink is created, visually connecting the source and target events. This edge is represented with a bold, directed arrow.
    \item To include an \textbf{XOR gate}, users can right-click on any node within the graph and select \textit{Add XOR Gate}. This action prompts users to enter the name of the gate. When submitted, the XOR gate is rendered and can be linked to primitive event nodes within the schema.
    \item Adding a new \textbf{entity} to the graph is straightforward as well. Users can right-click on any node and choose the \textit{add entity} option. They will then be prompted to provide the associated name, QNode, WikiData label, and description for the new entity. The provided information is immediately integrated into the schema's JSON structure. Figure \ref{add_entity} displays the prompt that is rendered when the user selects \textit{add entity}.
\end{itemize}

To access the existing entities in an uploaded schema, users can simply right-click on any node and select the \textit{view entities} option. This action presents a comprehensive list of entities, including their corresponding name, WikiData label, entity ID, and a list of events in which it participates. This overview facilitates the identification of potential additional instances of existing entities. \\

A collapsible \textbf{JSON editor} is included, allowing users to modify the schema's JSON representation directly. Upon uploading a schema, the JSON editor is automatically populated with the corresponding schema information. When changes are made by the JSON editor, these modifications are applied to the existing schema, resulting in an updated visual graph structure. This integration between the JSON editor and the visual representation ensures that any edits made to the schema properties are accurately reflected in the overall schema visualization. By incorporating a collapsible JSON editor, the interface offers users the flexibility to fine-tune and customize the schema according to their requirements. 

\section{Schema Instantiation}

After the first iteration of manual curation, schemas are passed to a schema-guided event prediction system, paired with an information extraction system for event forecasting. Based on a new set of news articles, the RESIN-pipeline \cite{wen2021resin} and RESIN-11 \cite{du2022resin} schema-guided event prediction system provided us with a list of matched and unmatched events covered by our event schemas. Because of the inherent noise from the evaluation step, the list of unmatched events becomes very long, with repeat occurrences of unmatched events and extraneous fine-grained events. To obtain a list of unmatched events that are of high importance and occur often, we first filter out fine-grained events like “go” and  “use”, as they are not in the scope of our high-level schema. Next, we perform a simple text encoding and matching step to identify event instances that may have been incorrectly identified as unmatched. Finally, we rank this filtered list of unmatched events based on frequency of occurrence. The final result of this process is a shortened list of unmatched events, where significant events tend to occur at a high frequency. 

Following this, a second iteration of manual curation is conducted with SCI 3.0 and the processed event list for reference. Included with the unmatched events are potential argument and entity names from the event prediction system, which are used to suggest arguments for events during curation. Table \ref{tab:eval} shows the approximate coverage of events prior to and after this second iteration of informed manual curation. By prioritizing frequently occurring concepts in news articles, we managed to double our event coverage across all curated schemas.

\section{Future Directions}
The most time-consuming process in schema curation has proven to be entity grounding due to the extensive effort required to search through WikiData. The Schema Curation Interface 3.0 is built with a compatible framework to the recent Human-in-the-loop schema system \cite{zhang2023humanintheloop}. By integrating this system into the SCI, curators will be able to prompt a dialog component with a few keywords to search candidate entities and relation Qnodes. Additionally, integrating the schema induction phase as well as the instantiation phase into one web application would streamline the process greatly.

This web interface has the potential for another valuable application: editing and curation of other graph-based semantic structures, like Abstract Meaning Representation (AMR) and Uniform Meaning Representation (UMR). This approach enables curators to add information via the user interface, bypassing the need to work directly with representation syntax. Our goal is to employ this methodology to streamline and accelerate existing data curation efforts in areas such as AMR and UMR.

Current efforts on the SCI are also directed towards making the tool more user-friendly. During the extensive curation process, it became clear that many of the sub-schema events could be reused across different schemas, even in different domains. We have the ability to separate hierarchical events from those which have no sub-events. If we were to create an entire library of sub-events, perhaps it would save the curator considerable time in recreating the same events which might be used elsewhere. This sub-schema library would be accessible to entire user groups for contribution and would be the building blocks of increasingly complex events.

\begin{table}[htbp]
    \centering
    \begin{tabular}{l|rr}
        \toprule
         & \textbf{Events} & \textbf{Participants} \\
        \midrule
        \textbf{Induced} & 376 & 957 \\
        \textbf{Manually Curated} & 377 & 604 \\
        \hline
        \textbf{Total} & 753 & 1561 \\
        \textbf{Increase (\%)} & 100 & 63 \\
        \bottomrule
    \end{tabular}
    \caption{Count of automatically induced events and participants relative to those manually added with the SCI 3.0.\protect\footnotemark}
    \label{tab:eval}
\end{table}
\footnotetext{\textit{Aggregate of chemical spill, coup, disease outbreak, general IED, radioactive spill, riot, and terrorist attack event schemas.}}

\section{Discussion}

A tool that can process complex event schemas into an interactive graphical representation enhances Information Extraction (IE), Data Mining, and NLP practices by clarifying relationships and patterns in large datasets. This pipeline can be used by a wide range of industries, such as healthcare, finance, and government agencies, to communicate complex data in an engaging and informative way. For example, government agencies and policymakers can use this tool to make more informed decisions and improve public policies. The development of this tool can have a significant societal impact by improving the accessibility and understanding of complex data, leading to better decision-making outcomes.

\section{Conclusion}
This work focuses on developing a Schema Curation Interface, which serves as a powerful web application for editing complex real-world event representations. By combining the capabilities of Large Language Models (LLMs) with human curation, our approach aims to enhance our understanding of natural language, cognition, and event conceptualization. The SCI 3.0 offers analysts a dynamic, user-friendly interface for real-time editing of event schemas, allowing the incorporation of chapter events, primitive events, participants, relations, and outlinks in a hierarchical graph. This facilitates a detailed mapping of event progressions and their temporal relationships.

\section*{Limitations}
As a manual task, the scalability of schema curation is left to the capabilities of the curator. Curation takes both time and attention to detail, meaning that the curation of a much larger schema library would be costly. The system may also struggle with the handling of ambiguous events. Curators have in the past found it difficult to decisively say when the start or end point of one event might be, leading to a measure of disagreement in scenarios.

\section*{Ethics Statement}
Both a limitation and an ethical consideration, the schema curation process is largely dependent on the domain expertise of the curator. This could inadvertently lead to misinformation or misrepresentation of events. Consequently, it is our duty as developers of these systems to provide analysts and curators with accurate schema inductions, and comprehensive materials in curating an event schema.


\nocite{*}
\section{Bibliographical References}\label{sec:reference}
\bibliographystyle{lrec-coling2024-natbib}
\bibliography{lrec-coling2024-example}

\label{lr:ref}
\bibliographystylelanguageresource{lrec-coling2024-natbib}
\bibliographylanguageresource{languageresource}

\appendix
\section{SDF Keyword Definitions}
\label{sec:SDF_definitions}

\textbf{Event Schema}

\begin{itemize}
    \item \texttt{@id} can be used to link JSON-LD elements, as each is uniquely assigned for every event, entity, participant, and relation.
    \item \texttt{sdfVersion} tracks the version of SDF used in each schema instance.
    \item \texttt{version} tracks the particular revision iteration of the event schema itself, useful when multiple curators are tasked with the same schema.
    \item \texttt{events} is the principle array of events, each with unique values.
\end{itemize}

\textbf{Events}

\begin{itemize}
    \item \texttt{@id} can be used to link JSON-LD elements.
    \item \texttt{name} is human-readable text to be displayed on the SCI 3.0.
    \item \texttt{description} is schema specific, describing the event in context to the schema, further than the Qnode is able.
    \item \texttt{wd\_node, wd\_label, and wd\_description} are the event grounding in Wikidata, this information can be found in the DWD.
    \item \texttt{isSchema} Boolean value returning true if the event has children sub-events.
    \item \texttt{repeatable} Boolean value returning true if the event can be repeated.
    \item \texttt{optional} Boolean value returning true if the event is optional.
    \item \texttt{children\_gate} string value of logical gate, can be OR or XOR for exclusive OR.
    \item \texttt{outlinks} an array of @id values, linking subsequent events.
    \item \texttt{participants} an array of participant objects.
    \item \texttt{children} an array of @id values, linking children events.
    \item \texttt{entities} an array of entity objects.
    \item \texttt{relations} an array of relation objects.
\end{itemize}

\textbf{Entities}

\begin{itemize}
    \item \texttt{@id} can be used to link JSON-LD elements.
    \item \texttt{name} is human-readable text to be displayed on the SCI 3.0.
    \item \texttt{wd\_node, wd\_label, and wd\_description} are the event grounding in Wikidata, this information can be found in the DWD.
\end{itemize}

\textbf{Participants}

\begin{itemize}
    \item \texttt{@id} can be used to link JSON-LD elements.
    \item \texttt{roleName} argument role found in DWD, provided by PropBank overlay.
    \item \texttt{entity} maps to the entity @id participating in the event.
\end{itemize}

\textbf{Relations}

\begin{itemize}
    \item \texttt{@id} can be used to link JSON-LD elements.
    \item \texttt{name} is human-readable text to be displayed on the SCI 3.0.
    \item \texttt{relationSubject} maps to the entity @id which is subject.
    \item \texttt{relationObject} maps to the entity @id which is object.
    \item \texttt{wd\_node, wd\_label, and wd\_description} are the event grounding in Wikidata, this information can be found in the DWD.
\end{itemize}

\textbf{Logic Gates}

\begin{itemize}
    \item \texttt{@id} can be used to link JSON-LD elements.
    \item \texttt{name} is human-readable text to be displayed on the SCI 3.0.
    \item \texttt{isSchema} Boolean value returning true if the event has children sub-events.
    \item \texttt{outlinks} an array of @id values, linking subsequent events.
    \item \texttt{children\_gate} string value of logical gate, can be OR or XOR for exclusive OR.
    \item \texttt{children} an array of @id values, linking children events.
    \item \texttt{optional} Boolean value returning true if the event is optional.
    \item \texttt{comment} defaults to the string "container node".
\end{itemize}

\section{Schema Curation Interface 3.0}

\begin{figure*}[b]
    \centering
    \includegraphics[width=1\textwidth]{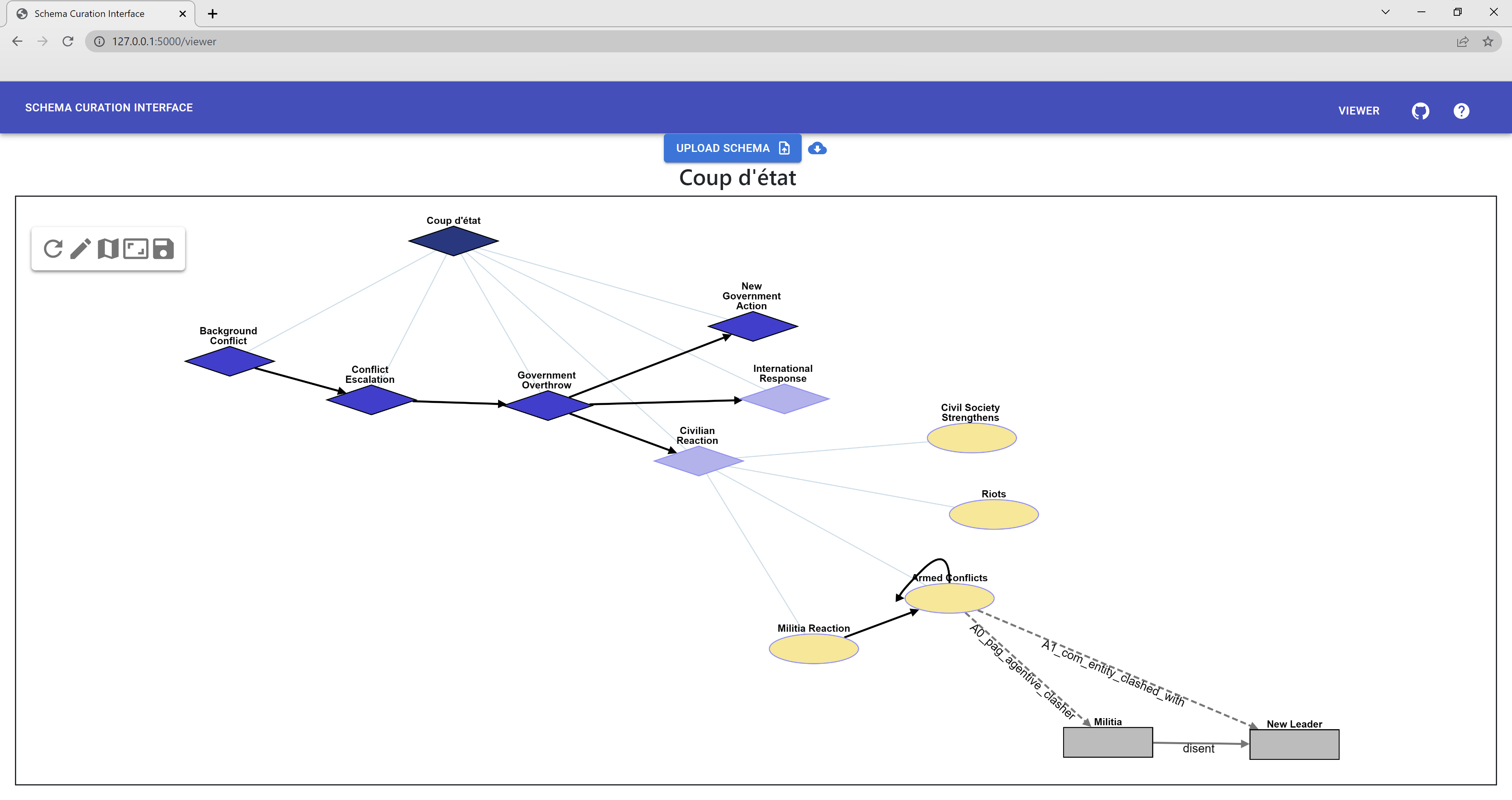}
    \caption{Overview of the rendered schema graph structure. Dark blue diamond elements represent chapter events. Light blue diamonds indicate optional chapter events. Similarly, the the yellow ovals in this image represent primitive events. The expanded primitive event includes participants, their role names, and relation to one another.}
    \label{graph_view_alt}
\end{figure*}
\label{sec:SCI_imgs}

\begin{figure*}[b]
    \centering
    \includegraphics[width=1\textwidth]{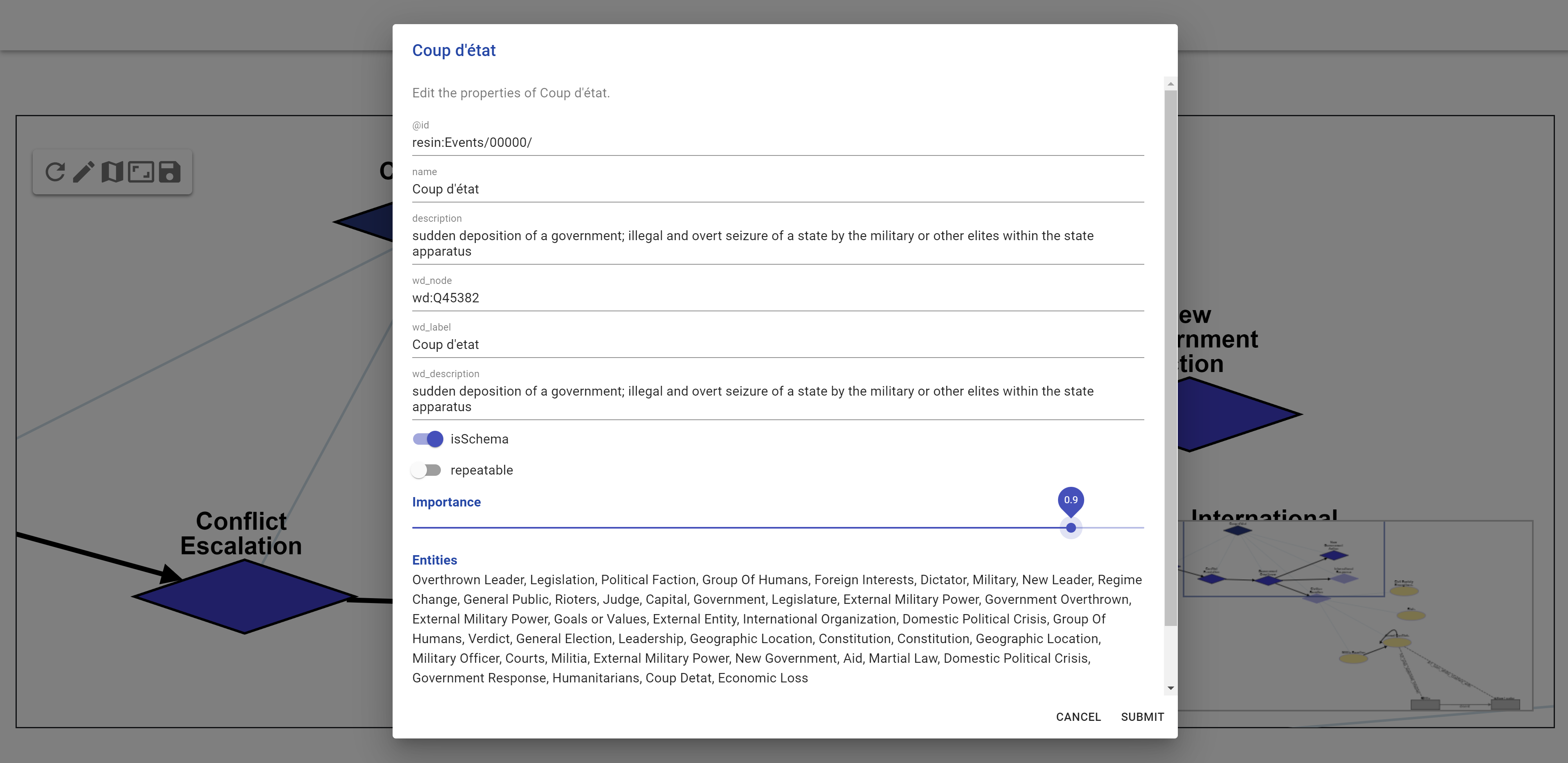}
    \caption{Example window for editing an event node. Allows users to signify event importance, edit event properties, and indicate optionality.}
    \label{edit}
\end{figure*}

\begin{figure*}[b]
    \centering
    \includegraphics[width=1\textwidth]{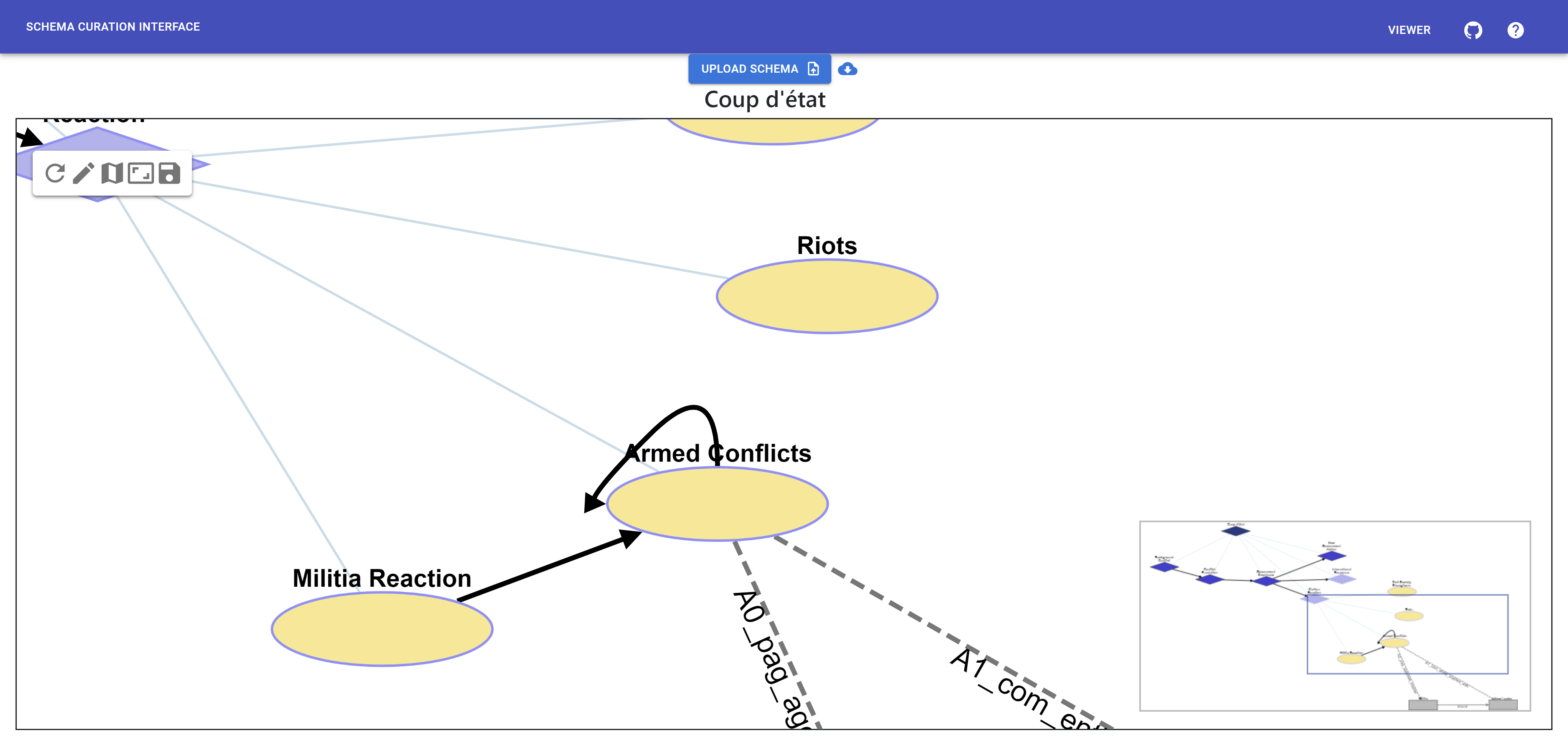}
    \caption{Illustration of the repeatability of event nodes, as well as the use of outlinks to indicate their temporal relationship. The navigator provides an overview of the entire graph structure in the bottom right corner, allowing users to orient themselves when viewing portions of the graph.}
    \label{navigator}
\end{figure*}
\end{document}